# Multi-layer Attention Mechanism for Speech Keyword Recognition

Ruisen Luo[1], Tianran Sun[1], Chen Wang[2], Miao Du[1], Zuodong Tang[1], Kai Zhou[1], and Xiaofeng Gong[1], and Xiaomei Yang[1,*]

[1]College of Electrical Engineering, Sichuan University, 24 South Section 1, One Ring Road, Chengdu, China 610065

[2]Department of Computer Science, Rutgers University -- New Brunswick, Piscataway, New Jersey, USA 08854

[*]The Corresponding Author, Xiaomei Yang, 398630203@qq.com

**Abstract— As an important part of speech recognition technology, automatic speech keyword recognition has been intensively studied in recent years. Such technology becomes especially pivotal under situations with limited infrastructures and computational resources, such as voice command recognition in vehicles and robot interaction. At present, the mainstream methods in automatic speech keyword recognition are based on long short-term memory (LSTM) networks with attention mechanism. However, due to inevitable information losses for the LSTM layer caused during feature extraction, the calculated attention weights are biased. In this paper, a novel approach, namely Multi-layer Attention Mechanism, is proposed to handle the inaccurate attention weights problem. The key idea is that, in addition to the conventional attention mechanism, information of layers prior to feature extraction and LSTM are introduced into attention weights calculations. Therefore, the attention weights are more accurate because the overall model can have more precise and focused areas. We conduct a comprehensive comparison and analysis on the keyword spotting performances on convolution neural network, bi-directional LSTM cyclic neural network, and cyclic neural network with the proposed attention mechanism on Google Speech Command datasets V2 datasets. Experimental results indicate favorable results for the proposed method and demonstrate the validity of the proposed method. The proposed multi-layer attention methods can be useful for other researches related to object spotting.**

***Keywords*—Automatic speech keyword recognition; dual-loop neural network with Attention mechanism; convolution neural network; bidirectional cyclic neural network.**

## Ⅰ Introduction

Traditional automatic speech recognition usually focusses on the recognition of an entire paragraph or passage of speech. Therefore, traditional speech recognition models usually need a vast amount of memory and computation resources. The huge sample size, together with the large memory cost and complex calculations occupied by the model, make traditional speech recognition hard to be proceeded when the resource is scarce. For example, in scenarios where there is no external computational resource and a microcontroller is the core of computing, traditional automatic speech recognition models become impossible to be used. As an alternative option, automatic speech keyword recognition technology has been paid more and more attention. In recent years, automatic speech keyword recognition technology has been applied to the aforementioned scenarios, and has produced positive outcomes on the performances.

Key challenges in the technology of speech keywords recognition exist mainly in the confusing noise from the high-variance speech characteristics without sematic meanings, such as different pronunciation colors and habits, speech cohesion, and ambiguous boundary between pronunciation units. Among the methods to overcome these problems, recurrent neural networks with attention mechanism has demonstrated promising capabilities. With explicit attention, recurrent neural networks are able to focus on the speech parts with clear sematic meaning and resist the impact of noise.

Conventional attention-based recurrent neural networks, mainly developed from the field of Natural Language Processing (NLP), use the output of the last layer to perform attention. However, in the application of Speech Recognition, the features are usually extracted as spectrogram or cepstrum, and the attention mechanism based on the last output will be insufficient. In this paper, to improve the performance of speech keyword recognition, a new attention model has been proposed. The new model considers attention mechanism on each level of the recurrent neural network, and leverages potential information especially from the feature layers.

Two datasets under four scenarios are tested based on the proposed multi-level attention recurrent neural network with multi-level output synergy. And for the purpose of comparison, the same experiment is performed on convolutional neural network, bidirectional recurrent neural network (Cyclic LSTM), and recurrent neural network with conventional attention model (Circulating LSTM). Experimental results show that the performance of the proposed model is consistently better than the compared counterparts. In addition, the proposed method in the paper has a better performance consistency: the performance of each test set is almost the same, and the reduction of individual keyword recognition rate is relatively low, which has a certain practical value.

### 1.1. Related Work

Using deep Neural Network and attention mechanism in Speech Recognition has been a long-time practice in the research community. In [1], the Gaussian mixture model and hidden Markov model are extended using the high-renowned feature extraction ability of deep neural network to supplement the deficiency of the original feature extraction model. By using 2000 hours of "SWB + Fisher" data set for training, the final error rate reached 13.8%. At the same time, Alex Graves also combines the Gaussian mixture hidden Markov model with LSTM unit of the cyclic neural network, and finds that the frame accuracy of the combined speech has been greatly improved in [2]. Igor Szoke simplifies the model, divides it into keyword model and background filling model, and calculates the maximum likelihood ratio in [3]. In [4], Ossama Abdel-Hamid uses convolutional neural network and weight sharing to recognize multi-keyword speech. Compared with traditional DNN keyword recognition model, YY improves the accuracy by 6%~10%. In [5], the experimental results of HaimSak et al. show that the combination of RNN and LSTM can reduce the frame rate in the case of a medium number of sample databases, and the recognition accuracy can be improved by about 5% compared with that of convolutional neural network. Golan Pundak et al. ([6]) constructed a Recurrent Highway Networks (RHN) deep LSTM cyclic neural network. This neural network uses "jump connection" to alleviate the problem of gradient explosion and gradient disappearance. It is also pointed out that the deep LSTM cyclic neural network constructed by RHN has a certain improvement compared with the baseline LSTM cyclic neural network. And finally, with respect to the attention mechanism, a notable work is by Jan K. Chorowski and his team ([7]), which proposed to introduce attention mechanisms into previous models to improve robustness to sequential feature memory in long sequence input. Furthermore, [8] conducted a work similar to this paper, which applied neural attention specifically for keyword recognition. However, it did not consider the multi-layer attention system proposed by this paper.

## Ⅱ Methods

### 2.1. Data pre-processing

#### 2.1.1 Mel-frequency Cepstral Coefficients

Following the standard process of speech recognition, speech samples need to be preprocessed before recognition. In our task of keyword spotting, since each audio segment in the database is a keyword recitation, and the length of each audio segment is 1 second, there is no need to use clipping processing. Therefore, we can choose to use Mel-frequency Cepstral Coefficients for audio preprocessing. Mel cepstrum is a spectrum representing short-term audio. Its principle is linear cosine conversion based on logarithmic spectrum represented by non-linear Mel scale, and Mel cepstrum coefficient is a set of coefficients used to establish Mel cepstrum. The extraction process of coefficients is mainly composed of the following four steps.

(1) Using framing, pre-emphasis and windowing to the original audio.

(2) Using Fourier transform to process audio.

(3) Reprocessing Fourier Transform Signals Using Mel Filter Banks.

(4) Using Discrete Cosine Conversion to Get MFCC.

Mel filter module includes mapping spectrum to Mel scale using triangular window function, and then multiplying the output of mapping with energy spectrum.

Using triangular window filter, the output of spectrum map mapped by Meyer scale is as follows:

$$H_m(k) = \begin{cases} \dfrac{k - f(m-1)}{f(m) - f(m-1)}, & f(m-1) \le k \le f(m) \\ 0, & others \\ \dfrac{f(m+1) - k}{f(m+1) - f(m)}, & f(m-1) \le k \le f(m) \end{cases} \quad (1)$$

After the output of filter is converted into energy spectrum, the Meier spectrum is obtained. The calculation formula is as follows:

$$\text{MELSPEC(M)} = \sum_{k=f(m-1)}^{f(m+1)} H_m(k) * |X(k)|^2 \quad (2)$$

Where $H_m(k)$ is the mask of Mel frequency cepstrum, $X(k)$ is is Fourier transform of signal, and $f(m)$ is is a triangular bandpass filter function.

## 2.2. Speech keyword recognition model based on traditional deep learning

In this section, the paper will discuss the major existed deep learning-based methods to perform speech recognition. These methods are also the methods-of-choice for the experimental comparisons with the proposed multi-layer attention method, of which the results are presented in section 2.3.

### 2.2.1 Automatic Speech Key Word Recognition Model Based on Convolutional Neural Network

We use convolutional neural networks as the front-end of complex neural networks to extract features and reduce the computational load of models. Figure 1 is a structure diagram of a classical convolutional neural network.

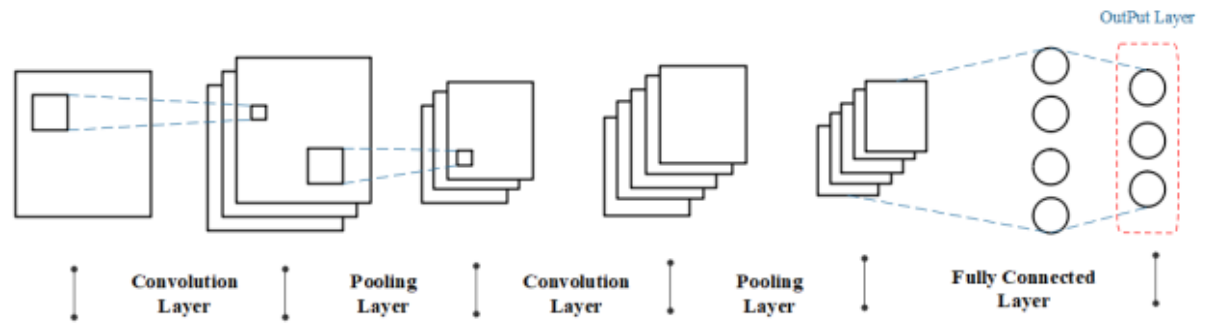

Fig.1 Structural Charts of Classical Convolutional Neural Networks [8]

In CNN-based speech keyword recognition, we first transform the original audio into Mel spectrum, and then send the sample transformed into two-dimensional picture to the convolution neural network for learning. The convolution neural network can extract the features of Mel spectrum, and finally classify it through the fully-connected layers.

The convolution neural network used in this paper consists of three layers of convolution layer (including pooling layer and Dropout) and three layers of full connection layer (as shown in Fig. 1). After each layer of convolution layer, batch standardization is used. The maximum pooling is used in the pooling layer, and the soft-max function is used as the activation function in the output layer of full connection layer.

### 2.2.2 Speech Keyword Recognition Based on Bidirectional LSTM Cyclic Neural Network

Compared with the CNN network mentioned above, the cyclic neural network has certain memory function, but it often forgets the information in the far ends. Therefore, LSTM units with three gate units are introduced on the basis of cyclic neural networks to filter and transmit information far away. Reduce the recognition error caused by the "forgetting" of the cyclic neural network.

When building a speech keyword recognition model using LSTM cyclic neural network, to use the convolution neural network introduced above to extract the features of the transformed two-dimensional image. Then the LSTM cyclic neural network is used to learn and map the output results. The basic flow chart is shown in Figure 2.

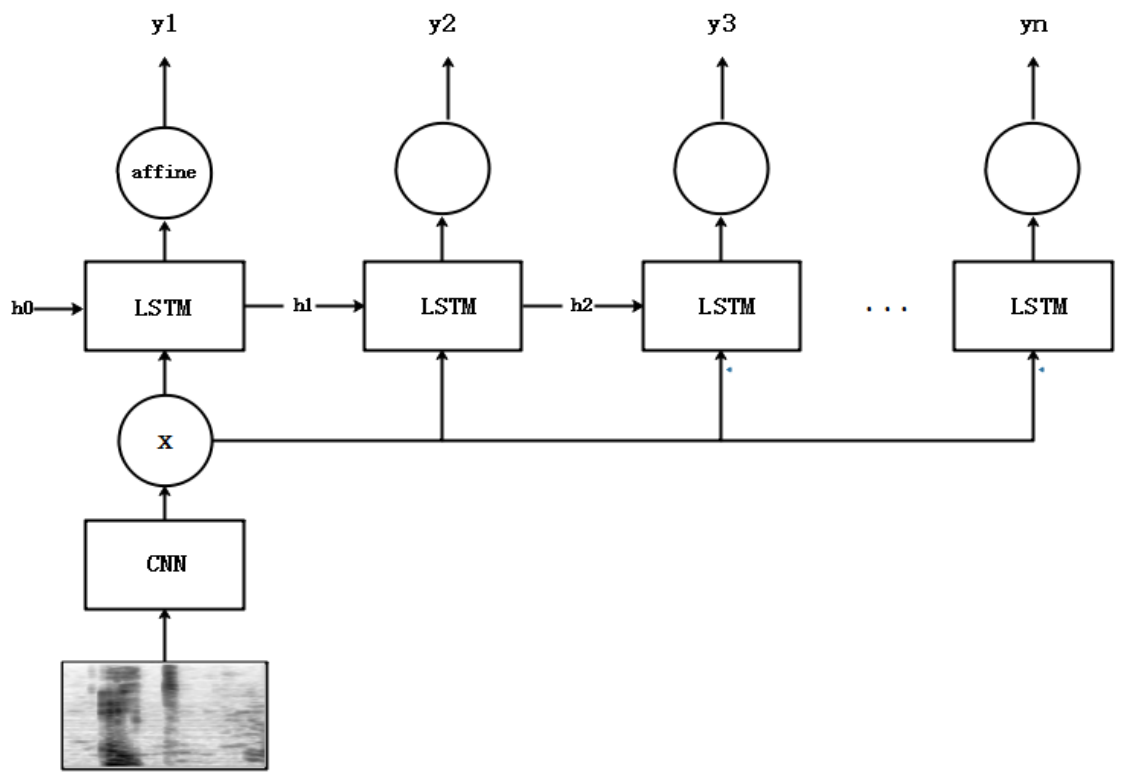

Fig.2 Basic Flow Chart of LSTM Cyclic Neural Network Model [9]

In this paper, the bi-directional LSTM cyclic neural network is combined with the convolution neural network. At first, the Meyer spectrum is extracted by two convolution layers. After dimensionality reduction, the output of the last convolution layer is input into the bi-directional LSTM cyclic neural network. The number of LSTM units is 64. Finally, the output layer of LSTM is transferred to the full connection layer for classification.

### 2.2.3 A Circulating Neural Network Model with Attention Mechanism

Speech keyword recognition model is usually built in the framework of Encoder-Decoder. It can be seen from Fig. 3 that the work of the Encoder-Decoder framework is to set two layers of network to be responsible for encoding the input and decoding it, respectively. For the vanilla encoder-decoder model, the output pays the same attention to each input after encoding, which could weight some noise as equally important to rich-sematic information and adversely affect the performance. Thus, the attention mechanism will make the system more attentive and devote more attention to the input it needs to pay attention to.

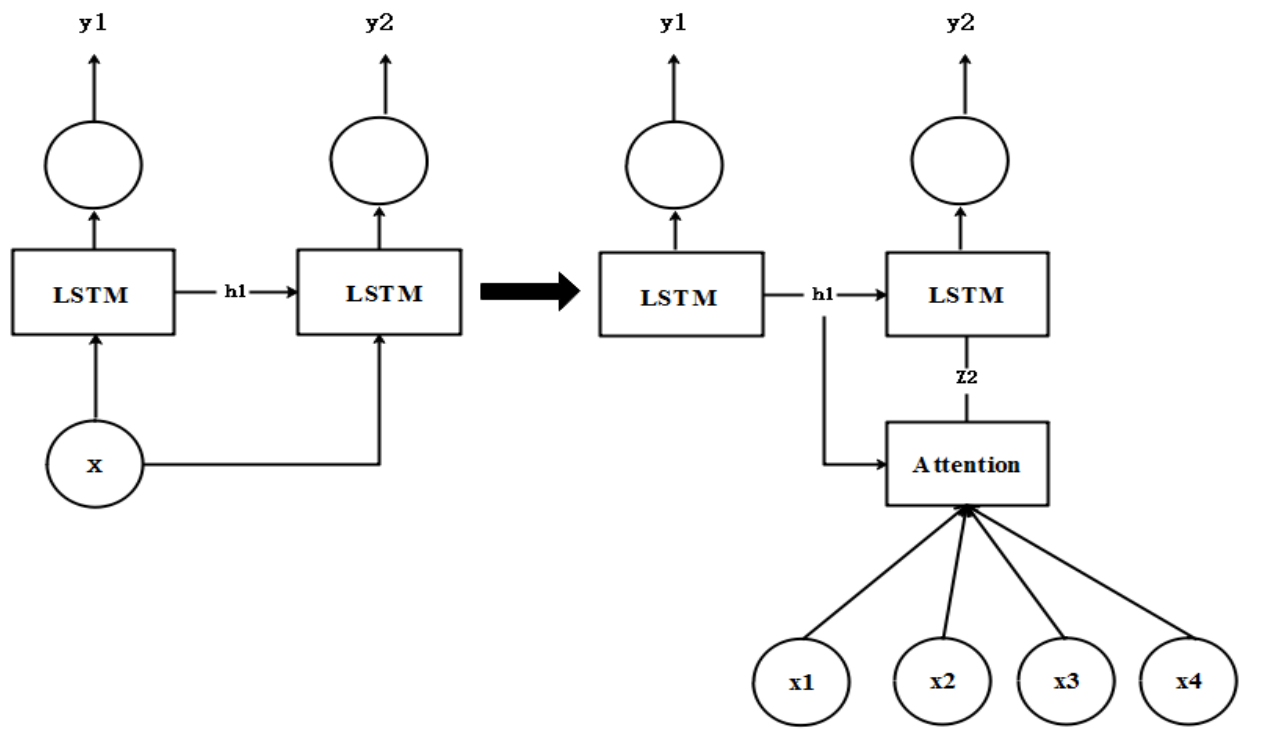

Fig.3 Basic Structural Chart of Circulating Neural Network with Attention Mechanism

In this paper, attention mechanism is introduced on the basis of bi-directional LSTM recurrent neural network. We trust that double-stacked LSTM recurrent neural network carries enough memory information.

### 2.3. Speech keyword recognition model with Multi-layer Attention Mechanism

Stemming from the application of machine translation, conventional circular neural network model with attention mechanism is based on extracting the output vector of the last LSTM layer, using dense layer projection and querying the vector to identify which part of the audio is most relevant. But when transferring to speech recognition, one significant problem will arise: the calculation of attention weights is solely based on output of the LSTM layer, and the input of LSTM is proceeded by Meyer Cepstrum (pre-processing) and extracted by convolution layer (feature extraction), which introduce information distortion inevitably. That is to say, certain important knowledge from previous layers are likely ignored, and the information in the attention mechanism might have biases.

Therefore, if the input of attention mechanism can be changed from only using the output layer of LSTM to the collaborative output of multiple layers in the overall process, it might be helpful for conquering the inaccurate attention weights problem. In this paper, a novel approach, namely Multi-layer Attention Mechanism, is proposed to handle the inaccurate attention weights problem. The key idea is that information of layers prior feature extraction and prior LSTM layer are also introduced into calculating attention weights. Therefore, the memory of the problem will be corrected and supplemented to improve the accuracy of keyword recognition.

The idea of the Multi-layer Attention model is visualized in Fig. 4 and a detailed description is shown in Fig. 5. Specifically, the information obtained by Meier cepstrum is dotted with the output layer of convolution neural network, and the feature is fused for the first time. Then the output of the first memory and the output of the LSTM intermediate layer are processed for the second memory operation, and the output of the second memory is obtained. Finally, the output of the second memory and the output of the last layer of LSTM are fused for the third time. The proposed multi-level attention mechanism memorize the parameters from each level to avoid missing crucial information from any layer of the system. Thus, it avoids the memory loss of some parameters caused by some reasons, which results in the low recognition rate of keywords.

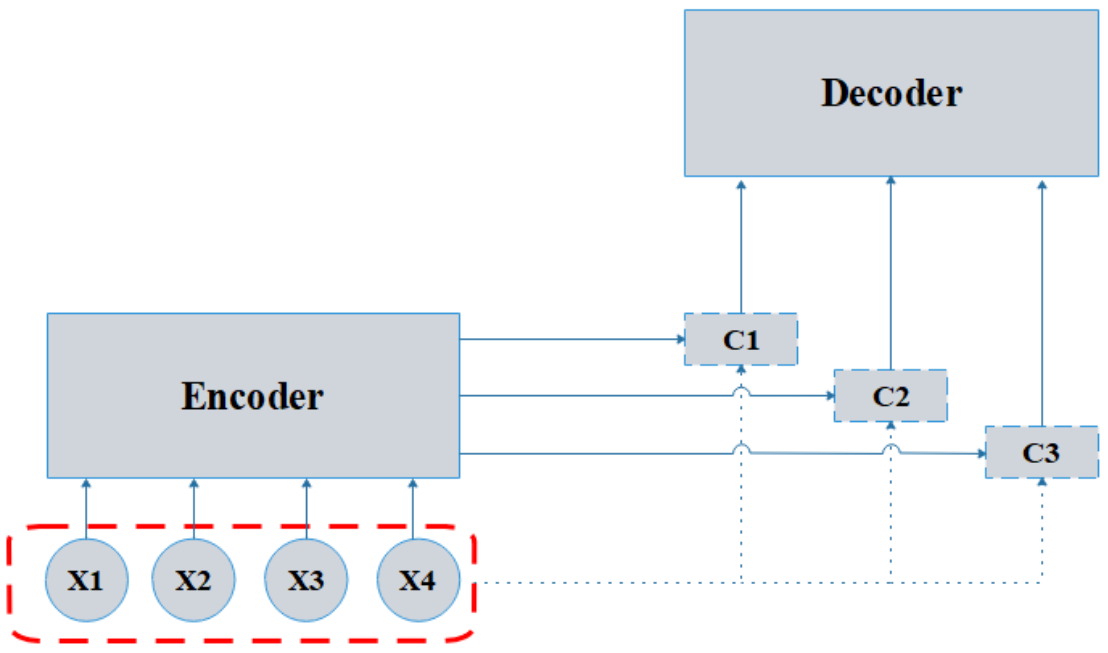

Fig.4 Multilayer Attention Model Framework []

It can be seen that the semantic encoding of input information at this time is not only related to the encoding mode, but also directly affected by the input. Therefore, the information carried is closer to the input information. Compared with the previous attention mechanism model, multi-level attention mechanism model will be affected by different levels of output synergy, and ultimately achieve better performance under multi-level synergy.

The keyword recognition model introduced in this paper is improved on the basis of the previous single-level attention mechanism. The original audio is transformed into a Meier spectrum, and the dimension-reduced output is first dotted with the output extracted by the convolution layer feature. The dot product is transferred through a full connection layer, then dot product with the output of the middle layer of LSTM layer, and the second full connection layer is transferred. Finally, the final attention parameters are obtained by dot product with the final layer of LSTM layer. After transferring the parameters to the output layer of LSTM, the memory information with attention weight assignment is obtained, which is eventually output to the full connection layer. The structure of the network model is shown in Fig. 5. The maximum number of cycles Epoch is 40 and the batch size is 64.

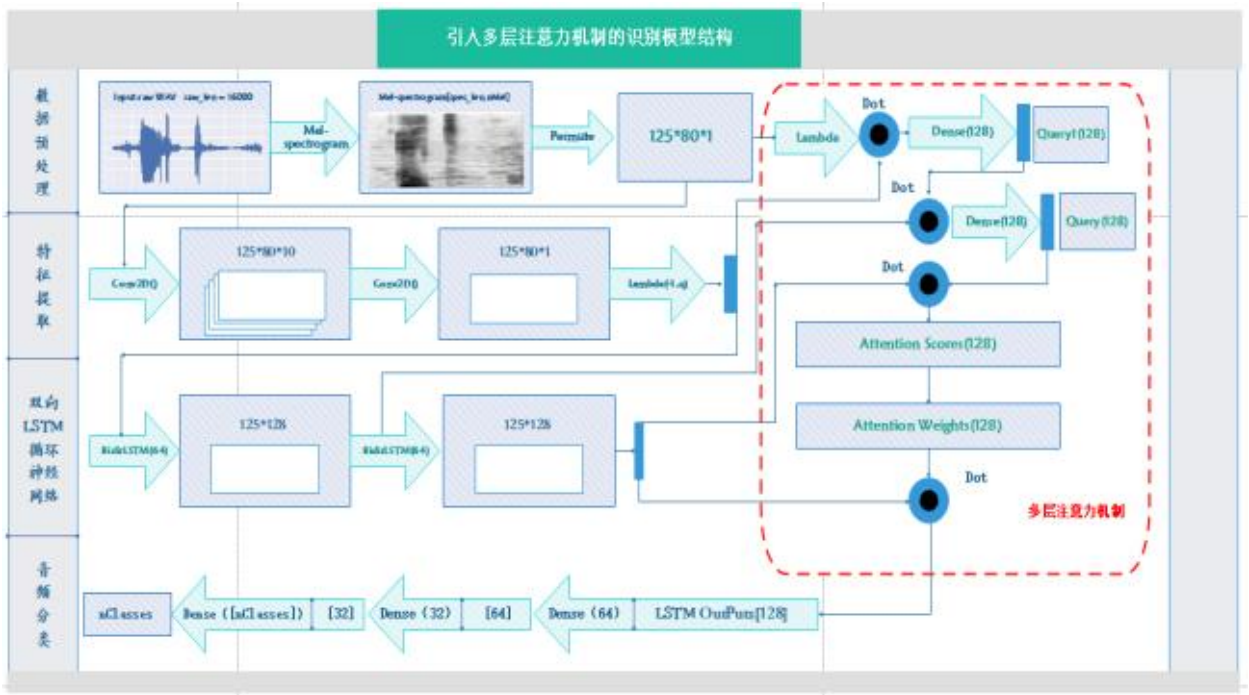

Fig.5 A Framework of Recognition Model with Multilayer Attention Mechanisms

## III Results and discussions

The training sample bank used in this paper is Google Speech Command datasets V2 with 20 keywords. We split the dataset with the following setups: the number of training and validation samples are 84849 and 9981, and the number of samples in the test set is 1105. The ratio between the 3 sets is about 8:1:1.

### 3.1. Mel-frequency Cepstral Coefficients

In the previous Fourier transform of audio, the accountant calculated the energy spectrum of audio on Fourier, and the output of the filter was transformed into the Mel spectrum. Log-Mel spectrum can be obtained by taking Mel spectrum as log.

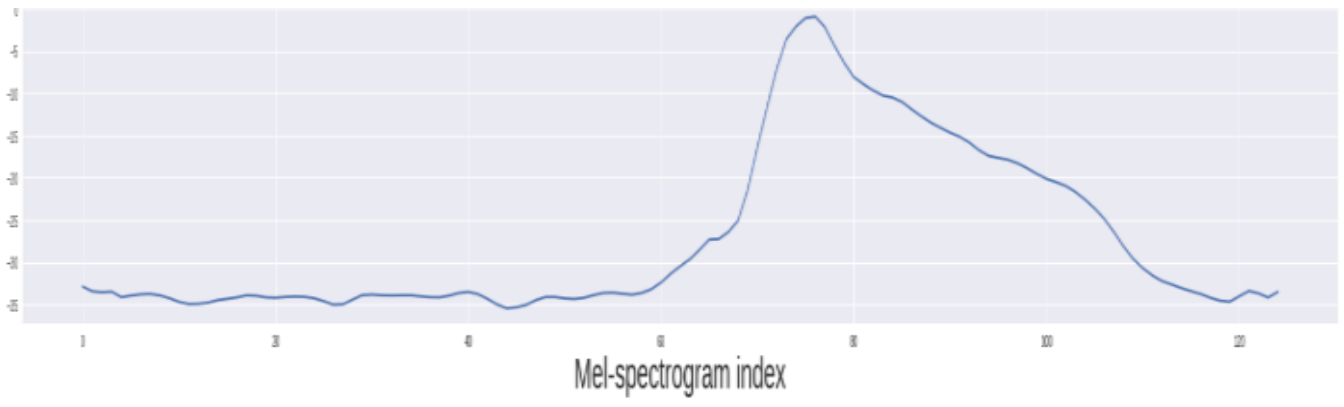

Fig.7 Log-Mel figure

Finally, the log-mel spectrogram is converted into discrete cosine to obtain the Mel cepstrum coefficients. The converted Mel spectrogram is shown in Fig.8.

Fig.8 Mel spectrum figure

After obtaining the Mel spectrum, we can put the spectrum into the built model for training.

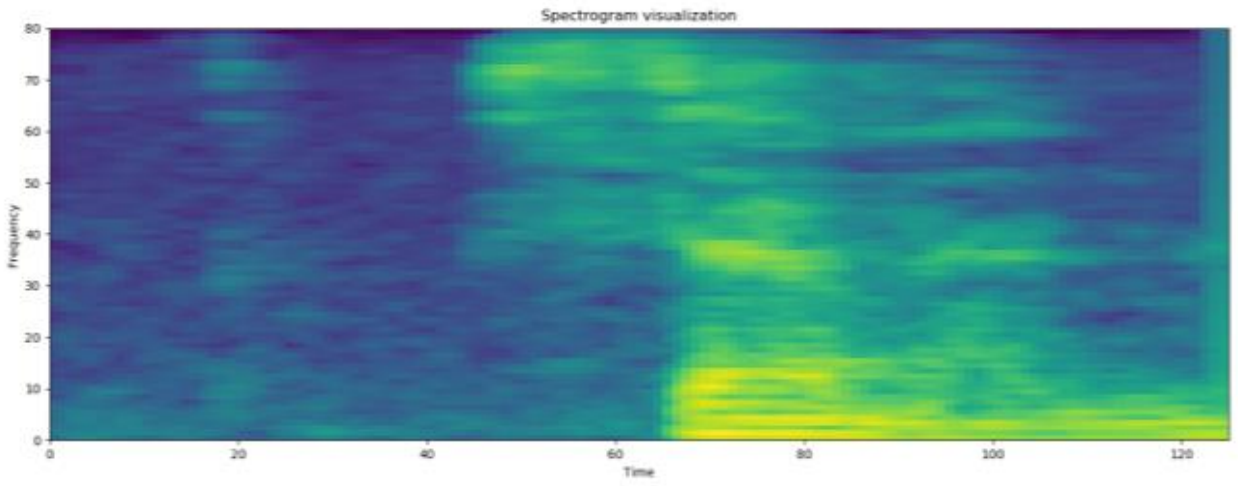

### 3.2. Recognition result

#### 3.2.1 Result on Existed Methods

The first model-of-comparison to be tested is the Convolutional Neural Network-based keyword recognition model. After the Convolutional Neural Network model is built, the pre-processed Meier spectrogram is imported, and the accuracy and loss function of the training set and the test set in the training process are stored with the parameters and optimizer unchanged. The drawing comparison is shown in Figure 9.

(a) Accuracy of training set and test set

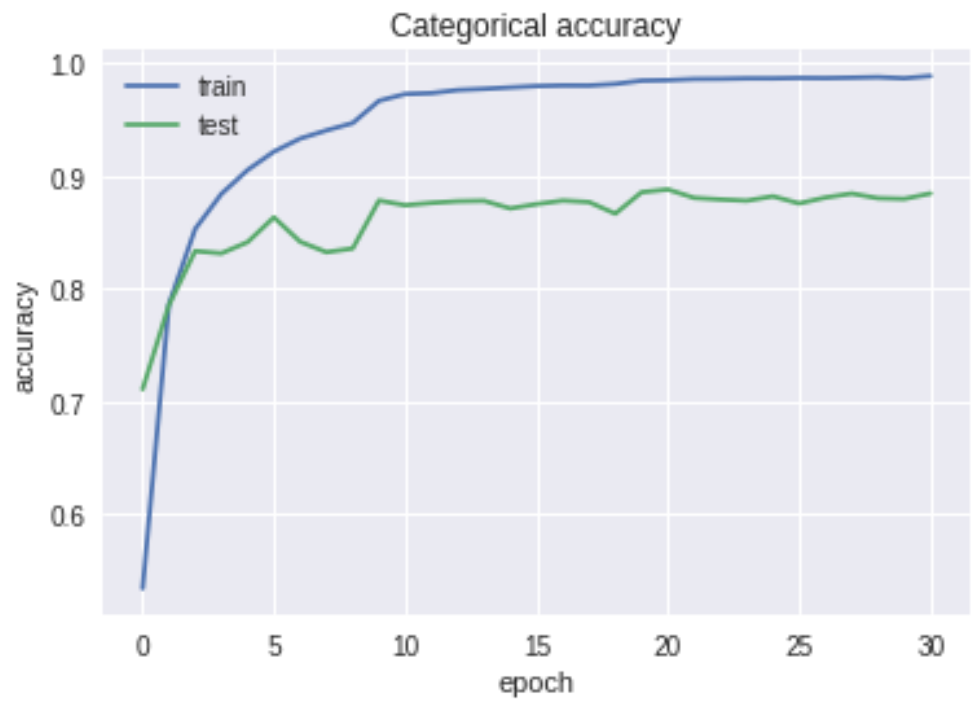

(b) Loss Function of training set and test set

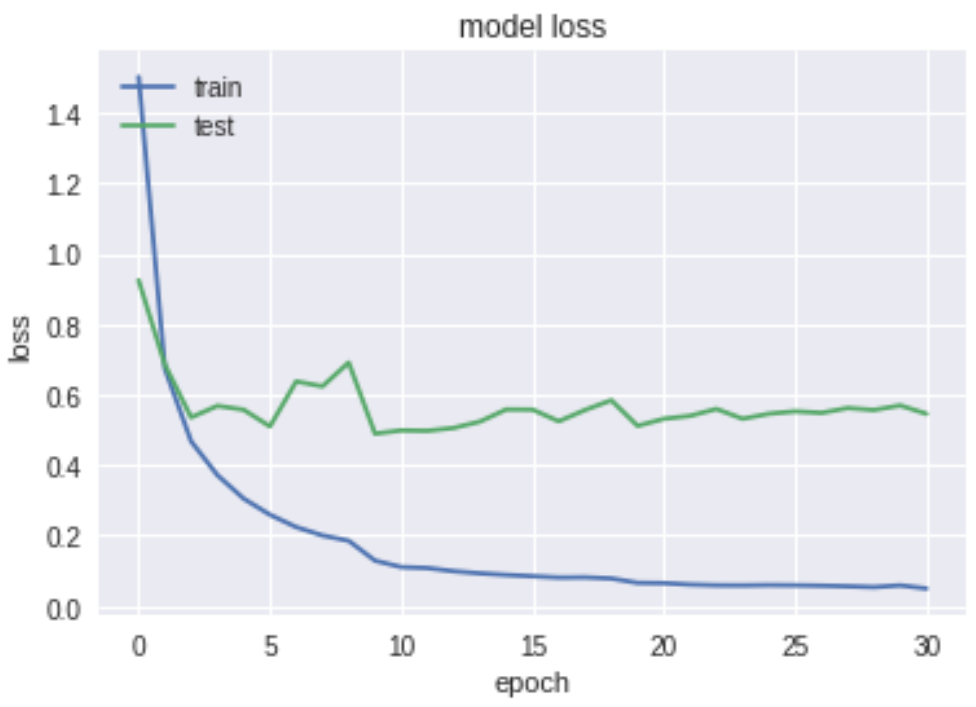

Fig.9 Loss Function and Accuracy Curve of Convolutional Neural Network-based Keyword Recognition

The second tested is the Bidirectional LSTM Cyclic

Preprint. Work in process.

Neural Network model. The accuracy and loss function of the training set and the test set in the training process are stored under the condition that the parameters and the optimizer are set unchanged, as shown in Figure 10.

The accuracy rate of bidirectional LSTM cyclic neural network on val-test set is 94.34%, which is much higher than that of the CNN-based method. However, excessive accumulation of LSTM units for assistant memory may cause degradation problems, which may lead to slow training speed or even stop training.

The third tested model, namely Circulating Neural Network with Attention Mechanism, could deal with the above problem and improve the recognition accuracy. In this model, the pre-processed data are imported into the Attention Mechanism model for training, and the accuracy and loss function of the training set and the test set in the training process are stored with the parameters and optimizer unchanged. The drawing comparison is shown in Figure 11.

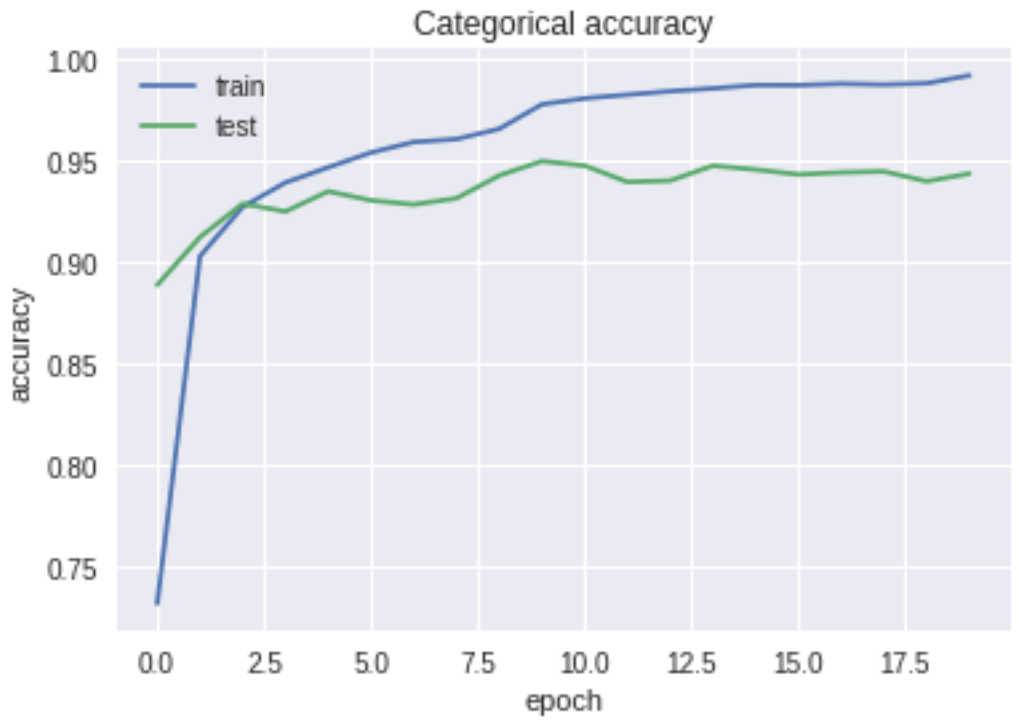

(a)Accuracy of training set and test set

The circular neural network with attention mechanism has accelerated the convergence, but the improvement on spotting accuracy is relatively insignificant. After a comprehensive analysis of the results on the test set, it is found that the LSTM circular neural network, with explicit attention or not, suffer most significantly from a low recognition rate for individual words. And in order to make the recognition rate of keyword model higher, this paper does it on the basis of the above model. Some improvements were made to improve the recognition rate of the model.

### 3.2.2 Result on Multi-layer Attention LSTM

Like the previous network, in order to avoid the program falling into gradient explosion or making meaningless calculation without gradient, the decision of early stopping is added to the training, and the best-preserved model is selected. The model is optimized by using Adam optimizer with learning rate attenuation. It can be seen that the accuracy of the test set at the end of the first Epoch is 0.87933, the loss function of the test set is 0.4332, the running time of each step is 149 ms, and the running time of each cycle is 198 seconds; the accuracy of the test set in the eleventh cycle is 0.95030, and the loss function of the test set is 0.2291. In the 21st cycle, the test set has not exceeded the accuracy of 11 cycles, so the trigger lift is triggered. Early_stop, end the loop.

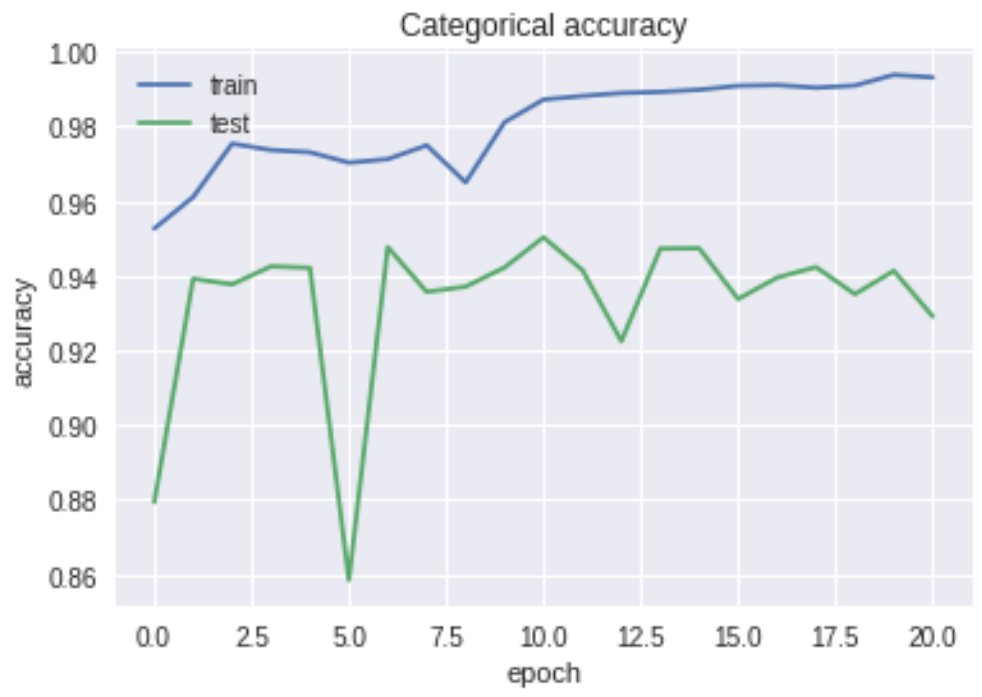

(a)Accuracy of training set and test set

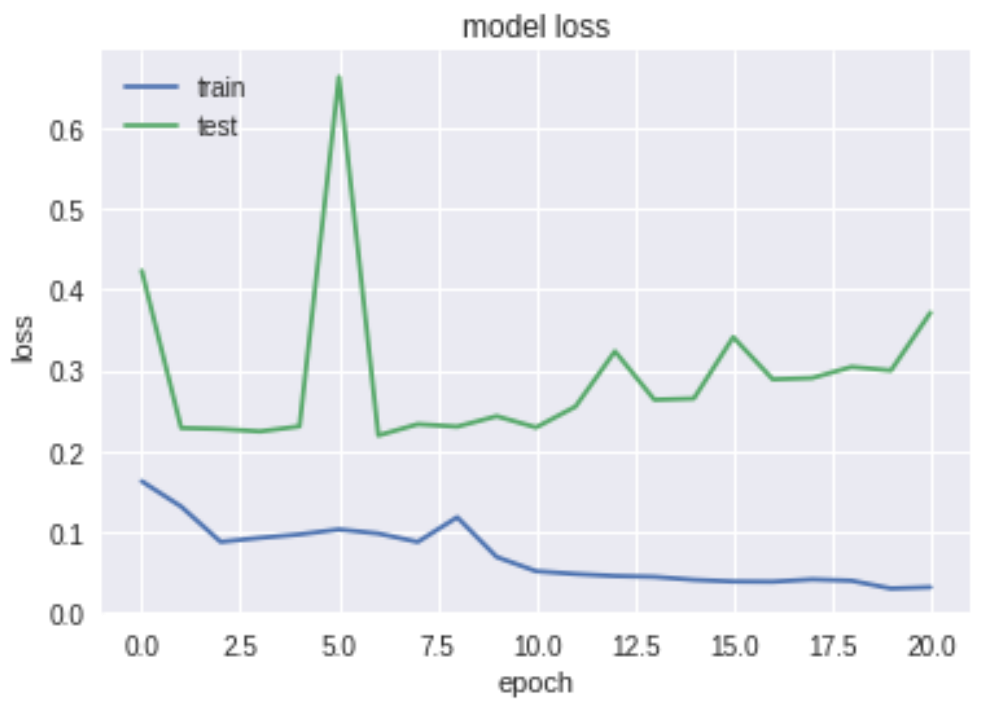

(b)Test Set and Training Set Loss Function

Fig.9 Loss function and accuracy curve of keyword recognition model with multi-layer attention mechanism

After the model was built, the number of samples was 8484849, and there were 20 keywords. Similarly, the maximum number of cycles Epoch was 40, and the batch_size was 64. Like the previous network, in order to avoid the program falling into gradient explosion or making meaningless calculation without gradient, the decision of early stopping is added to the training, and the best-preserved model is selected. The model is optimized by using Adam optimizer and learning rate attenuation. The accuracy and loss functions of the training set and test set of each cycle in the training process are stored and compared as follows.

Through the later verification, it is proved that the speech recognition model with attention feature has good performance on different test sets, and the recognition rate without keywords is very low, which is in line with our assumption.

To further analyze the insight of the results based on multi-layer attention, the histogram of spotting accuracy of the 20 keywords is plotted as Figure 11. From the results of figure 11, the recognition accuracy of each word is above 93%, except the 'down' command has a lower recognition rate of 89.2% (slightly lower than 90%). The overall recognition rate is 95.07%, which is 0.7% higher than that of the previous model with attention mechanism, and the accuracy is further improved.

In order to reduce the phenomenon that the recognition rate of individual keywords is too low in the above models, this paper also proposes a cyclic neural network model with attention features to solve the above problems. The histogram of the recognition accuracy of 20 keywords using the cyclic neural network model with attention features is as follows:

The recognition rate of the model is 93.72%. Although the recognition rate is lower than that of the cyclic neural network model and the multi-layer attention mechanism model, we can see from the above figure that there is basically no keyword recognition rate is relatively low, and there is no keyword recognition rate less than 90%. In addition, we use other sample databases and test sets to validate the data set, and find that the performance of the model is relatively stable, and the performance on other data sets is better than that on this data set.

## IV Conclusion

This paper presents a speech keyword recognition model with multi-layer attention mechanism. In the case of Google Speech Command datasets V2 dataset, the audio signal is preprocessed to obtain the Mel cepstrum coefficient. Through comprehensive comparisons with convolutional neural network, bi-directional LSTM cyclic neural network and cyclic neural network with attention mechanism, it is shown that the proposed multi-layer attention mechanism could improve the performance of Keyword Spotting. Furthermore, with a comprehensive analysis on the spotting accuracy of individual keywords, it could be found that the proposed method could achieve high accuracy for all but one keyword. The proposed method improves the state-of-the-art keyword spotting performance, and the multi-layer attention mechanism makes an algorithmic contribution to the field.

## References

Preprint. Work in process.